\documentclass[letterpaper]{article} 
\usepackage{aaai2026}  
\usepackage{times}  
\usepackage{helvet}  
\usepackage{courier}  
\usepackage[hyphens]{url}  
\usepackage{graphicx} 
\usepackage{natbib}  
\usepackage{caption} 
\usepackage{algorithm}
\usepackage{algorithmic}

\usepackage{newfloat}
\usepackage{listings}
\DeclareCaptionStyle{ruled}{labelfont=normalfont,labelsep=colon,strut=off} 
\floatstyle{ruled}
\newfloat{listing}{tb}{lst}{}
\floatname{listing}{Listing}
\title{Toward Gaze Target Detection in Young Autistic Children}
\author {
    Shijian Deng\textsuperscript{\rm 1},
    Erin E. Kosloski\textsuperscript{\rm 2},
    Siva Sai Nagender Vasireddy\textsuperscript{\rm 1},
    Jia Li\textsuperscript{\rm 1},
    Randi Sierra Sherwood\textsuperscript{\rm 2},
    Feroz Mohamed Hatha\textsuperscript{\rm 1},
    Siddhi Patel\textsuperscript{\rm 2},
    Pamela R. Rollins\textsuperscript{\rm 2},
    Yapeng Tian\textsuperscript{\rm 1}
}
\affiliations {
    \textsuperscript{\rm 1}Department of Computer Science, The University of Texas at Dallas\\
    \textsuperscript{\rm 2}School of Behavioral and Brain Sciences, The University of Texas at Dallas\\
    \{shijian.deng, erin.kosloski, sivasainagender.vasireddy, jia.li, randi.sherwood, feroz.hatha, siddhi.patel, rollins, yapeng.tian\}@utdallas.edu
}

\usepackage{bibentry}
\usepackage{amsmath}
\usepackage{amssymb}
\usepackage{booktabs}
\usepackage{hyphenat}
\usepackage[table]{xcolor}

\definecolor{myblue}{RGB}{0,87,120}    
\definecolor{rowgray}{gray}{0.93}      

\begin{document}

\maketitle

\begin{abstract}
The automatic detection of gaze targets in autistic children through artificial intelligence can be impactful, especially for those who lack access to a sufficient number of professionals to improve their quality of life. 
This paper introduces a new, real-world AI application for gaze target detection in autistic children, which predicts a child's point of gaze from an activity image. This task is foundational for building automated systems that can measure joint attention—a core challenge in Autism Spectrum Disorder (ASD). To facilitate the study of this challenging application, we collected the first-ever Autism Gaze Target (AGT) dataset. We further propose a novel Socially Aware Coarse-to-Fine (SACF) gaze detection framework that explicitly leverages the social context of a scene to overcome the class imbalance common in autism datasets—a consequence of autistic children's tendency to show reduced gaze to faces. 
It utilizes a two-pathway architecture with expert models specialized in social and non-social gaze, guided by a context-awareness gate module. The results of our comprehensive experiments demonstrate that our framework achieves new state-of-the-art performance for gaze target detection in this population, significantly outperforming existing methods, especially on the critical minority class of face-directed gaze.
\end{abstract}

\begin{links}
    \link{Code}{https://github.com/ShijianDeng/AGT}
\end{links}

\section{Introduction}
\label{intro}

\begin{figure}[t]
\centering
\includegraphics[width=1\columnwidth]{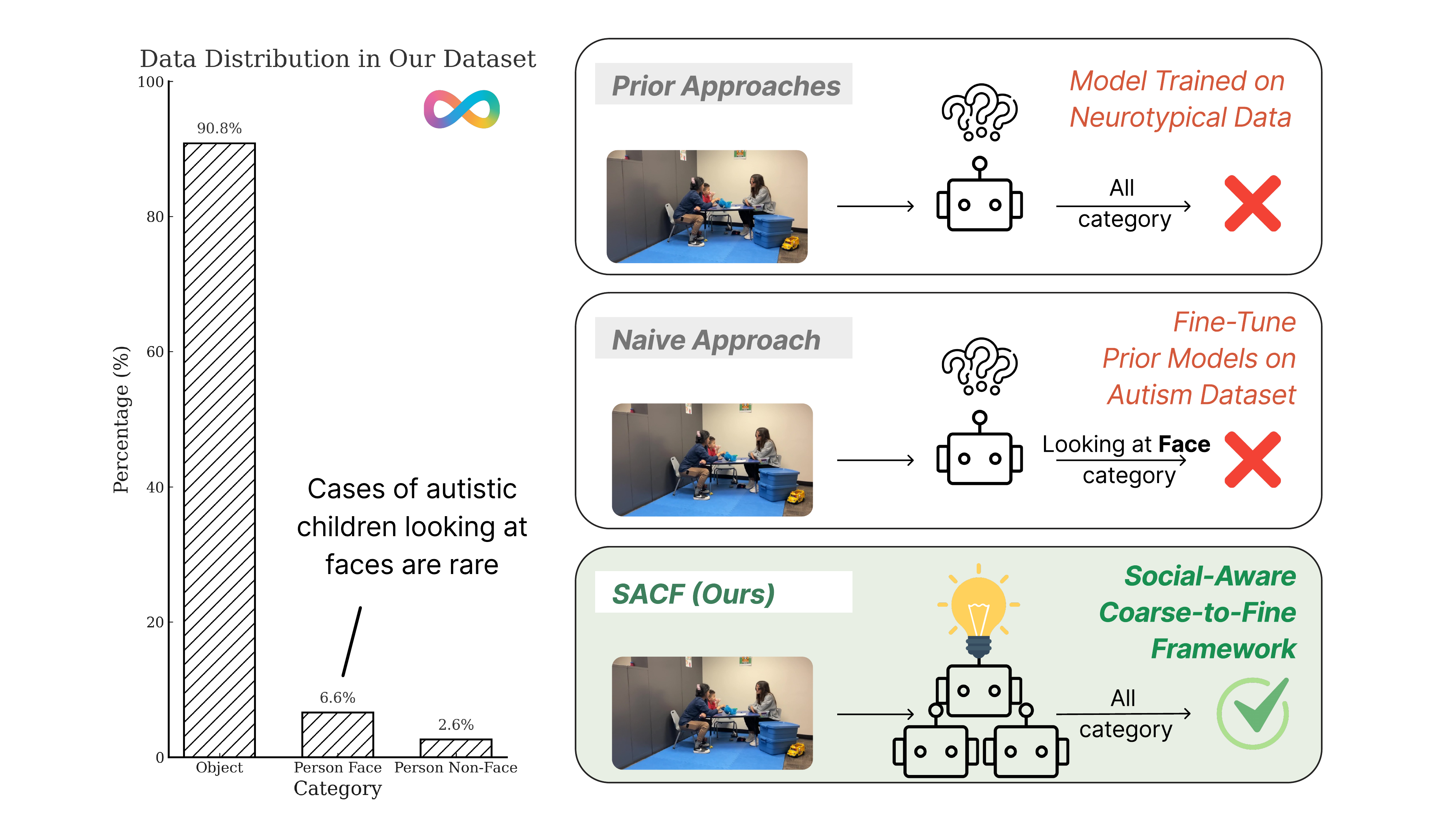}
\caption{For gaze target detection in young autistic children: (a) Existing off-the-shelf models trained on neurotypical data cannot handle autism scenarios well because autistic individuals can have a different behavior distribution in the same social environment compared to neurotypical individuals, such as avoiding eye contact or not responding to verbal interaction. (b) Models directly fine-tuned on an autism dataset still struggle with the rare case of an autistic child looking at a person's face. (c) Our Socially Aware Coarse-to-Fine framework uses an MLLM as a router to dynamically utilize a Socially Aware Gaze Expert model and a Socially Agnostic Gaze Expert to address these issues.}
\label{fig:teaser}
\end{figure}

Autism Spectrum Disorder (ASD) is a neurodevelopmental condition characterized by challenges in social interaction, communication, and the presence of restricted or repetitive behaviors~\cite{guha2014diagnostic}. The CDC estimates that about 1 in 31 eight-year-olds has been identified with ASD~\cite{shaw2025prevalence}. A core feature often observed in young autistic children is a difference in social attention, particularly in initiating and responding to joint attention \cite{bruinsma2004joint,jones2013attention,zwaigenbaum2005behavioral}. 
Joint attention refers to the shared focus of two individuals on an object and is a critical milestone in early social development. Its assessment is a cornerstone of early autism diagnosis and intervention, yet it is a process that is labor-intensive, requires highly trained professionals, and is difficult to scale~\cite{luyster2009autism}.

Artificial intelligence (AI) holds immense promise to augment the capabilities of clinicians by providing objective, scalable, and accessible tools for behavioral assessment. One of the most fundamental technical challenges in building such a tool is automatic gaze target detection: can an AI model determine where an autistic child is looking within a naturalistic setting? If successful, such a system could quantify instances of joint attention, track progress over time, and provide valuable feedback to caregivers and clinicians.

However, existing research on gaze target detection has focused almost exclusively on neurotypical adults~\cite{nips15_recasens, chong2020detecting}, and more recently, neurotypical children~\cite{tafasca2023childplay}. These models are typically trained on datasets where gaze toward faces is common. When applied to autistic children, their performance degrades significantly. This is due to a fundamental difference in behavior: autistic children tend to look at faces less frequently than their neurotypical peers~\cite{klin2002visual,chawarska2013decreased}.

To address the bias introduced by neurotypical datasets, we collected the first autism-specific dataset for gaze target detection. Nevertheless, autism related gaze behaviors also introduce a severe class imbalance. Gaze toward faces, although clinically significant, is underrepresented in the data. As a result, the trained models tend to predict non-social targets and could miss critical moments of social engagement.

To overcome this challenge, we developed a sophisticated, Socially Aware framework to handle the low-frequency gaze-to-face events in the data. Our socially aware framework leverages the large-scale prior knowledge learned by multi-modal large language models (MLLMs)~\cite{Qwen2.5-VL} to first coarsely recognize the social context, then process and route the input to fine-grained gaze target detection expert models that can better handle such scenarios. Our approach helps us achieve new state-of-the-art performance on gaze target detection in autistic children, especially in socially-related cases, which is crucial for downstream tasks such as joint attention detection. 
 
Our contributions can be summarized as follows: 1) We tackle a new, challenging, and socially impactful task: \textbf{Autism Gaze Target Detection}, which takes an image containing an autistic child and outputs the gaze target of that child; 2) We collect and annotate the first-ever \textbf{Autism Gaze Target (AGT)} dataset, a real-world dataset of autistic children in semi-structured play sessions, designed specifically for this task. It contains 16,582 annotated frames and reflects the natural gaze distribution of this population; 3) We propose a \textbf{Socially Aware Coarse-to-Fine (SACF)} gaze detection framework. Instead of a single, monolithic model, SACF first assesses the social context of the scene and then deploys specialized expert pathways for social (face-directed) and non-social gaze. This approach directly confronts the class imbalance problem and improves performance on the crucial, yet rare, social gaze events in the autistic population; 4) Our experiments show that SACF significantly outperforms state-of-the-art baselines, demonstrating its potential as a foundational component for AI-driven tools aimed at supporting the autism research community.

\section{Related Work}
\label{related_work}

\subsection{AI for Autism}
The application of AI and computer vision to autism research has gained significant traction. Early work focused on analyzing static images or videos to classify behaviors. 
Researchers have developed models to detect autism-related behaviors \cite{rajagopalan2013self,pandey2020guided,negin2021vision,wei2023vision,ribeiro2023stimming,deng2024language,deng2024hear}, analyze gesture and gait \cite{zahan2023human}, and classify movements \cite{singh2024video}. Some efforts have shifted towards quantifying attention dynamics. For example, \cite{jiang2017learning} uses a DNN-based model with eye-tracking data in free image reviewing. Others have developed an attention-based autism screening model that uses photo-taking and image reviewing for ASD \cite{chen2019attention}. 
Building on this line of research, we focus on gaze target detection in young autistic children in naturalistic social settings. Rather than analyzing gaze in isolation, we aim to understand how a child visually engages with the surrounding environment, including people and objects. Gaze target detection in this context offers a socially meaningful lens into early social attention development and supports the creation of tools that may assist in developmental assessment and intervention.

\subsection{Gaze Target Detection}
Gaze target detection aims to identify the object or person someone is looking at in an image or video. 
Early methods focused on narrow areas \cite{fathi2012social}. The advent of natural settings revolutionized the field. The GazeFollow model by \citet{nips15_recasens} was a seminal work, introducing a dataset that was not limited to a specific domain. This in-the-wild approach spurred further research. Subsequent research introduced datasets leveraging temporal information from video \cite{chong2020detecting}.
However, these prominent models and datasets are based on neurotypical adults~\cite{fang2021dual,gupta2022modular,bao2022escnet,jin2022depth}. The recently proposed Childplay dataset \cite{tafasca2023childplay} made strides by focusing on neurotypical children in play scenarios. While valuable, their data distribution still reflects a high frequency of social gaze. No existing work has systematically studied or addressed the unique challenges of gaze detection in autistic children, where the data distribution is inherently and dramatically skewed away from social targets. Although previous research~\cite{tafasca2023childplay} emphasizes the importance of the direction of autism gaze target detection, and \citet{chong2020detecting} provided preliminary evidence for automating gaze-based joint attention detection for toddlers, our work is the first to systematically tackle this specific, challenging, and clinically vital domain on a large-scale dataset.

\section{Autism Gaze Target Dataset}
\label{dataset}
To facilitate the new research, we constructed the Autism Gaze Target (AGT) dataset. This dataset was collected and annotated under Institutional Review Board (IRB) approval with informed consent from all participants' legal guardians.

\begin{figure}[t]
\centering
\includegraphics[width=0.75\columnwidth]{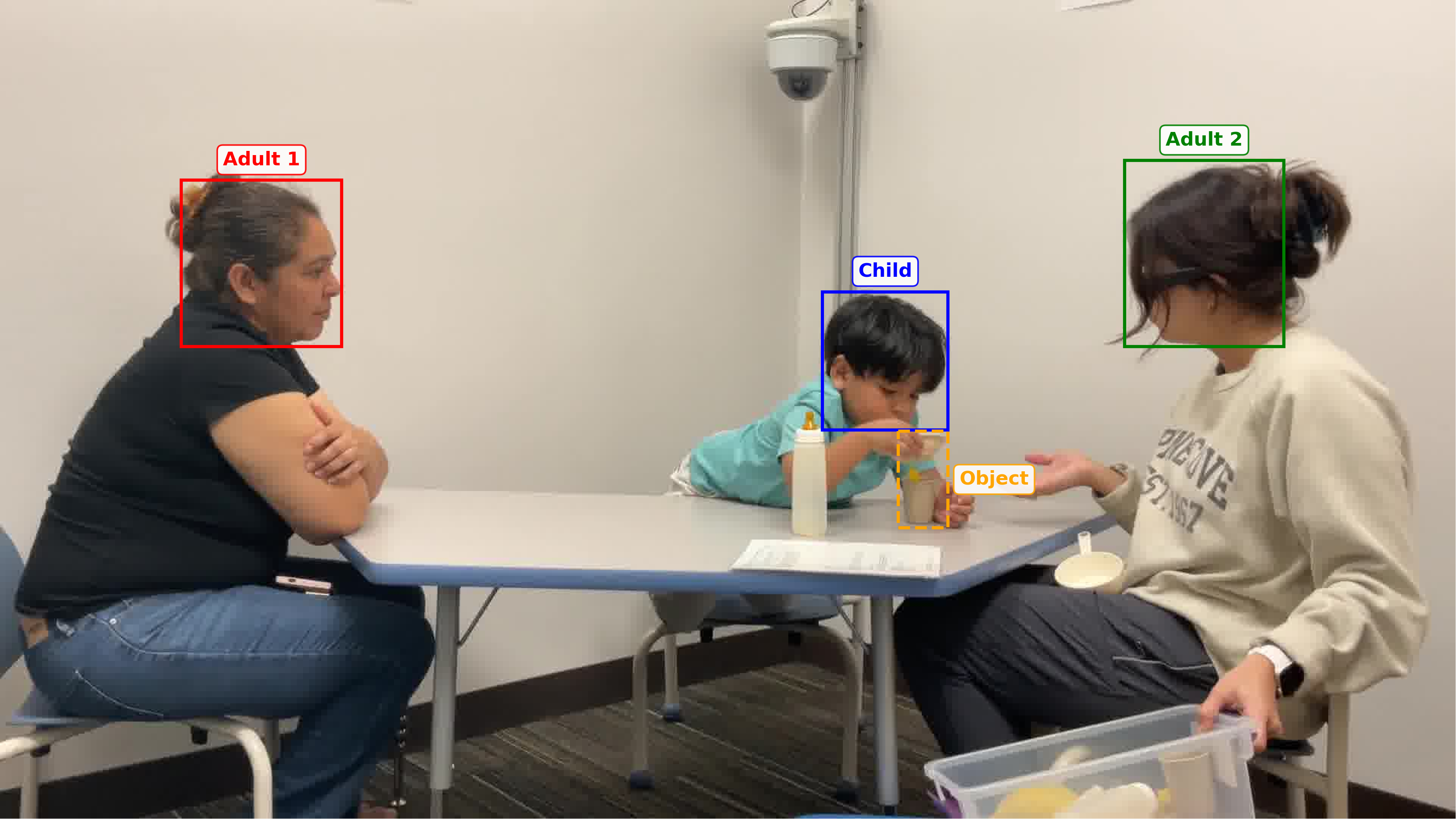}
\caption{Example from our Autism Gaze Target (AGT) dataset. The child's head is bounded by a blue box. The clinician's and parent's faces are marked with green and red boxes, respectively. The ground-truth target (a toy) is highlighted with a dashed orange box.}
\label{fig:dataset_examples}
\end{figure}

\subsection{Data Collection}
We extracted nearly 20,000 video frames from 59 ethically sourced video recordings, which are part of the Autism Corpus built by autism experts on our team~\cite{rollins2021mutual,rollins2023reexamining}.
Video recordings of autistic children were collected during administration of the Communication and Symbolic Behavior Scales-Developmental Profile~\cite{wetherby2002communication} by a research clinician, as part of two broader studies.

\begin{figure}[t]
\centering
\includegraphics[width=0.7\columnwidth]{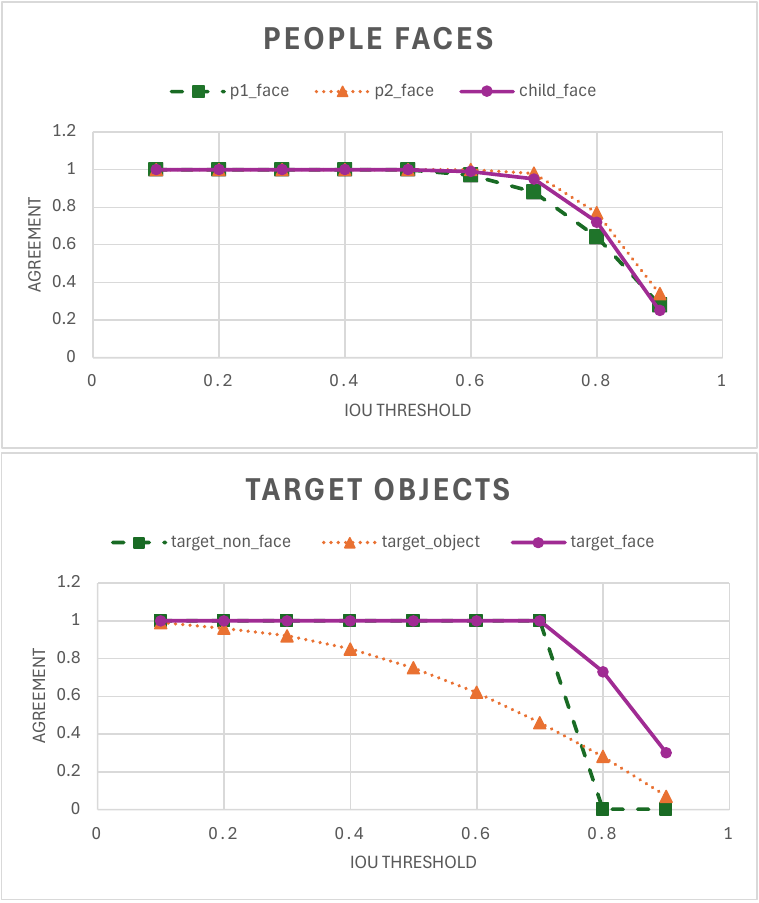}
\caption{Annotation agreement. Most gaze targets are objects. Agreement remains high and drops slowly as the IoU Threshold increases, indicating strong consistency between annotators.}
\label{fig:agreement}
\end{figure}

\begin{table}[t]
  \centering
  \small
  \renewcommand{\arraystretch}{1.25}
  \setlength{\tabcolsep}{12pt}
  \begin{tabular}{
    >{\columncolor{myblue}\color{white}\bfseries}c
    c c c}
    \rowcolor{myblue}
      & \textcolor{white}{\textbf{Object}}
      & \textcolor{white}{\textbf{Face}}
      & \textcolor{white}{\textbf{Non\hyp{}face}} \\
    Object        & \cellcolor{rowgray} 499 & \cellcolor{rowgray} 4  & \cellcolor{rowgray} 12 \\
    Face          & \cellcolor{rowgray} 4   & \cellcolor{rowgray} 27 & \cellcolor{rowgray} 1  \\
    Non\hyp{}face & \cellcolor{rowgray} 12  & \cellcolor{rowgray} 1  & \cellcolor{rowgray} 1  \\
  \end{tabular}
  \caption{Confusion matrix of looking-target categories also shows high agreement between annotators.}
  \label{tab:confusion_matrix}
\end{table}

\subsection{Data Annotation}
From the recorded videos, we sampled frames at 1 frame per second (fps) and grouped every five consecutive frames into a clip. This approach reduces temporal redundancy while preserving meaningful attentional shifts. For each sampled frame, a team of trained annotators used the Computer Vision Annotation Tool~\cite{cvat} to manually label social and non-social gaze targets. The annotation process involved the following steps:

Annotators used bounding boxes to mark the positions of:
\begin{enumerate}
    \item \textbf{Child Face}: A bounding box for the child's face.
    \item \textbf{Adult Faces}: Bounding boxes for all adults' faces in the scene. These were the clinician's face or the parent's face.
    \item \textbf{Ground-Truth Target}: Annotators were instructed to carefully examine the child’s head pose and eye direction to identify the target the child was most likely looking at. Each gaze target was labeled as one of three categories: a target object, a target face, or a target located on a person but not the face, with the corresponding visual region marked. A “Noninclusive” category was used when the gaze is directed outside the frame or at no specific target.
\end{enumerate}
Each frame was annotated by at least one annotator, resulting in a total of 16,582 fully annotated video frames after filtering out frames without a gaze target. One example is illustrated in Figure~\ref{fig:dataset_examples}.
To ensure annotation quality, we randomly selected 800 frames for double annotation, where two annotators independently labeled the same frames.
Figure~\ref{fig:agreement} presents the inter-annotator agreement based on the Intersection over Union (IoU) criterion, computed as:

\begin{equation}
\text{agreement} \;=\;
\begin{cases}
  1 & \text{if IoU} > \text{threshold},\\[4pt]
  0 & \text{otherwise}.
\end{cases}
\end{equation}
We also generated a category confusion matrix (Table \ref{tab:confusion_matrix}), which yields a Cohen's Kappa score~\cite{kvaalseth1989note} of 0.757, indicating substantial agreement.

\begin{table}[t]
\centering
\begin{tabular}{lrr}
\toprule
\textbf{Gaze Target Class} & \textbf{Frame Count} & \textbf{Percentage} \\
\midrule
Face (More-social) & 1,088 & 6.6\% \\
Not Face (Less-social) & 15,494 & 93.4\% \\
\midrule
\textbf{Total} & \textbf{16,582} & \textbf{100.0\%} \\
\bottomrule
\end{tabular}
\caption{Gaze target distribution in the Autism Gaze Target (AGT) dataset. The severe imbalance, with `Face' as the minority class, poses a significant challenge for standard models.}
\label{tab:dataset_stats}
\end{table}

\subsection{Data Statistics}
The final AGT dataset reflects the expected gaze patterns in the young autistic population. As shown in Table \ref{tab:dataset_stats}, the distribution of gaze targets is highly imbalanced. Gaze is directed towards the adult's face in only 6.6\% of the frames. The majority of gaze is not directed towards face (93.4\%), but instead toward other targets, including objects and person-non-face regions.
This imbalance means standard models will likely struggle with this data, as the most clinically relevant class (`Face') is a minority class. 
We split the data into training (9874 frames), validation (3344 frames), and test (3364 frames) sets to ensure that our models generalize to unseen participants.

\section{Method}
\label{method}
In this work, we aim to accurately predict a child's gaze target, with a particular emphasis on correctly identifying the rare but crucial instances of social gaze towards a face.

\subsection{Problem Formulation}
The task of gaze target detection aims to identify the specific location a person is looking at within a visual scene. In this paper, we formalize this task for the unique context of young autistic children in naturalistic settings. Given an input image $I \in \mathbb{R}^{H \times W \times 3}$ and the bounding box of the child's head $B_{head}$, the primary objective is to predict the 2D coordinates of the gaze target, $(x_t, y_t)$, within the image frame.

Beyond spatial localization, a critical aspect of this problem involves the semantic interpretation of the target. Let $C = \{\text{Face, Not Face}\}$ represent the set of possible target categories. In addition to predicting the gaze coordinates $\mathbf{p}_t = (x_t, y_t)$, the model must also assign a semantic label $c_t \in C$ to the identified target. Therefore, the problem is to build a comprehensive framework $\mathcal{F}$ that maps the input image $I$ and head bounding box $B_{head}$ to a tuple containing both the predicted location and its class:
\begin{equation}
    (\mathbf{p}_t, c_t) = \mathcal{F}(I, B_{head})
\end{equation}
Successfully learning this function is essential for downstream clinical applications. The primary challenge is to optimize for this joint objective in a domain characterized by severe data imbalance, where the prior probability of the `Face' class is significantly lower than that of the `Not Face' class, i.e., $P(c_t = \text{Face}) \ll P(c_t = \text{Not Face})$.

\subsection{Preliminary}
To address the localization challenge defined above, modern methods typically train a model to produce a probabilistic gaze heatmap, from which the final coordinates are derived. 
Our framework builds upon such a state-of-the-art model with prior knowledge augmented by MLLMs.

\paragraph{Gaze Target Detection Model.}
For fine-grained gaze target localization, we use \textbf{GazeLLE} \cite{ryan2025gaze}, a state-of-the-art architecture designed for efficiency and accuracy. It revolutionizes previous methods by leveraging a single, frozen, pre-trained image encoder, $\Phi_{enc}$ (e.g., DINOv2~\cite{oquab2023dinov2}), to extract a feature representation, $F_{scene} \in \mathbb{R}^{N \times D}$, from the entire scene image $I$.

Instead of a secondary network for the head, GazeLLE injects the subject's location directly into the feature space using a person-specific positional prompt, $E_{pos}$, derived from the head bounding box $B_{head}$. This combined representation is then processed by a lightweight transformer decoder, $\mathcal Dec_{gaze}$, to produce the final gaze heatmap, $M_{gaze} \in [0, 1]^{H' \times W'}$. The process can be summarized as:
\begin{equation}
    M_{gaze} = \mathcal Dec_{gaze}(\Phi_{enc}(I) + E_{pos}(B_{head}); \theta_{dec})
\end{equation}
where only the decoder parameters, $\theta_{dec}$, are trained. The model is optimized using a ground-truth heatmap, $\hat{M}_{gaze}$, which is generated by applying a Gaussian blur centered on the ground-truth target coordinates. The training objective is to minimize the pixel-wise Binary Cross-Entropy (BCE) loss between the predicted and ground-truth heatmaps:
\begin{align}
    \mathcal{L}_{gaze} = -\frac{1}{H'W'} \sum_{i,j} \Big( & \hat{M}_{gaze}^{(i,j)} \log(M_{gaze}^{(i,j)}) \nonumber \\
    & + (1 - \hat{M}_{gaze}^{(i,j)}) \log(1 - M_{gaze}^{(i,j)}) \Big)
\end{align}
This architecture's effectiveness and streamlined design make it an ideal base model for our problem. We also use the recent \textbf{Sharingan} model \cite{tafasca2024sharingan} as another base model.

\paragraph{Multimodal Large Language Model}
To tackle the semantic interpretation aspect of the problem, we leverage \textbf{Qwen2.5-VL-7B-Instruct} \cite{Qwen2.5-VL}, a small MLLM. We employ it as a Social Context Awareness (SCA) module to provide a high-level, coarse understanding of the scene. Given the image $I$ and a textual prompt $T$, the MLLM estimates a social context score $s$ corresponding to the probability that the child is looking at a face:
\begin{equation}
    s = P_{SCA}(V_{answer} = \textit{looking at face} | I, T; \theta_{MLLM})
\end{equation}
This score allows our framework to dynamically adapt its strategy based on the semantic potential of the scene.

\begin{figure*}[t]
\centering
\includegraphics[width=0.95\textwidth]{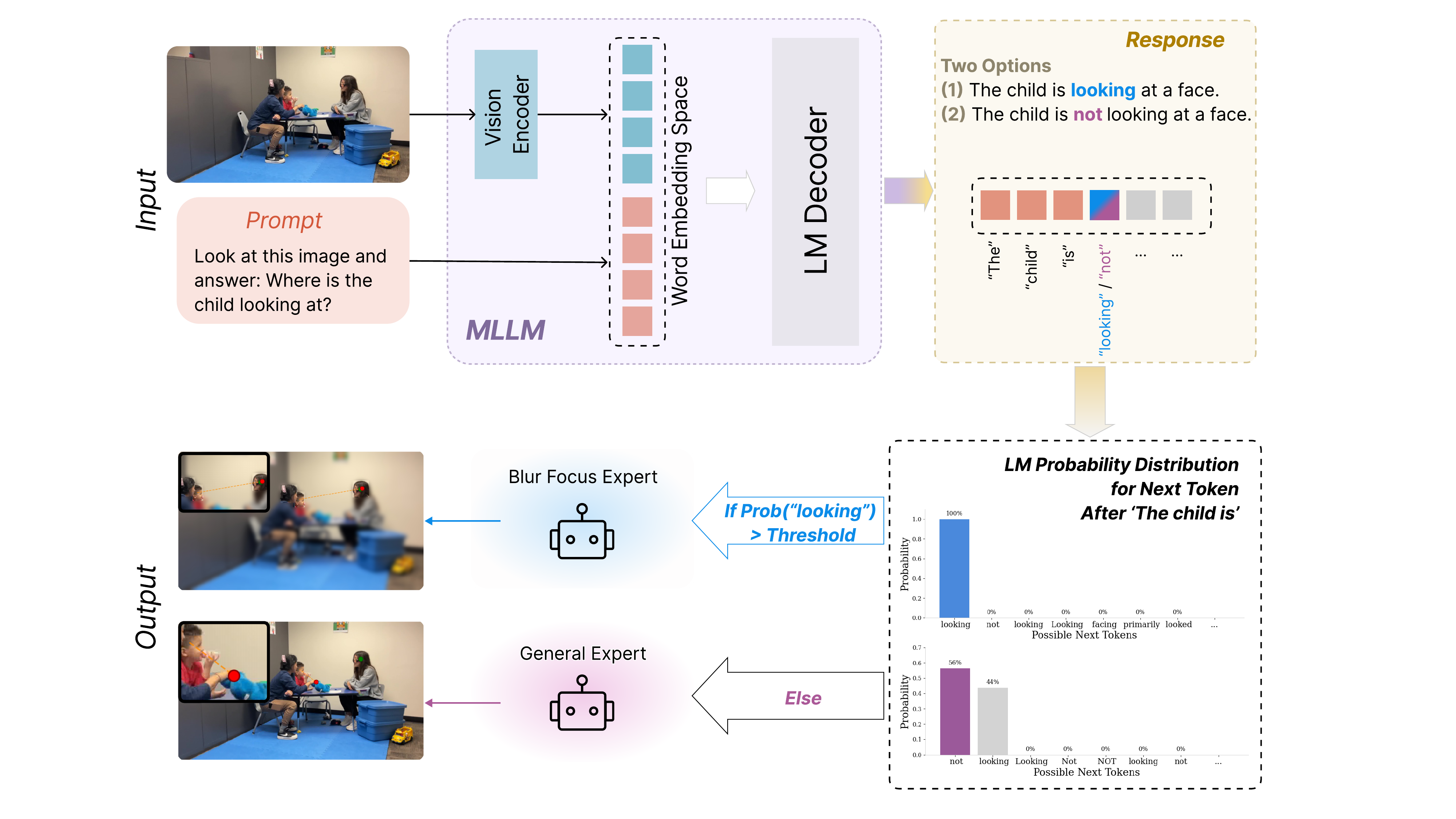}
\caption{The architecture of our proposed Socially Aware Coarse-to-Fine (SACF) gaze detection framework. An input image is first processed by the Social Context Awareness (SCA) module to generate a social context score $s$. This score is used by a gate to dynamically assign the input image to two specialized expert pathways: a Socially Aware Gaze Expert ($Ex_{aware}$, Blur Focus) and a Socially Agnostic Gaze Expert ($Ex_{agnostic}$, General). The final prediction is derived from the routed expert outputs, allowing the framework to adapt its specialization based on the scene's social context.}
\label{fig:model_architecture}
\end{figure*}

\subsection{Socially Aware Coarse-to-Fine Framework}
The central challenge in gaze target detection for autistic children stems from the severe class imbalance in the data. Standard models fail on this task because their optimization objectives are inherently compromised by this distribution, where the probability of the clinically critical `Face' class is dwarfed by the `Not Face' class. Such models inevitably learn to prioritize the majority class, achieving low average error by effectively ignoring the rare but vital social gaze.

To counteract this bias, our core idea is that a ``divide and conquer'' strategy that decomposes the problem into coarse-grained semantic understanding and fine-grained specialized localization would outperform a single, monolithic model. This motivation guides our proposed \textbf{Socially Aware Coarse-to-Fine (SACF)} framework, $\mathcal{F}_{SACF}$. As illustrated in Figure \ref{fig:model_architecture}, the framework first assesses the scene's social context before routing it to an expert model fine-tuned for that specific context.

\paragraph{Social Context Awareness (SCA) Module.}
To enable a ``divide and conquer'' approach, the framework must first answer: \textit{What is the potential looking target in this scene?} A powerful MLLM, with its vast pre-trained knowledge of visual scenes, can serve as an effective estimator for this high-level task. We therefore implement the SCA module using a fine-tuned Qwen2.5-VL model. It is trained on a binary proxy task derived from our data, $\mathcal{D}_{SCA} = \{I_i\}_{i=1}^N$, to estimate the social context score $s$, which represents the likelihood that the child is looking at an adult's face.
The framework uses the social context score $s$ from the SCA module to gate the two pathways via a simple threshold, $\tau$. This determines the coarse class prediction $c_{coarse}$:
\begin{equation}
    c_{coarse} =
    \begin{cases}
        1 & \text{if } s \ge \tau \\
        0 & \text{if } s < \tau
    \end{cases}
\end{equation}
Here, $1$ denotes face and $0$ denotes not face. This provides the coarse assessment of the scene's social context needed to guide the subsequent fine-grained analysis.

\paragraph{Two-Pathway Gated Experts.}
Given a coarse classification from the SCA, the next challenge is to perform accurate localization. A single fine-grained model would still be influenced by the data imbalance. 
We therefore hypothesize that training two separate experts, $Ex_{aware}$ and $Ex_{agnostic}$, with distinct and specialized objectives will be more effective. Each expert model, based on the GazeLLE architecture, maps an image and head box to a gaze heatmap, $Map = Ex(I, B_{head}; \theta)$.

\subparagraph{Socially Aware Gaze Expert ($Ex_{aware}$).}
The rationale for this expert is to create a specialist for the case when the social context is clear. We train it on an augmented dataset, $\mathcal{D}_{aug}$, where we apply a strong Gaussian blur, $G_{blur}$, to irrelevant regions of an image $I$ when the target is certain. This encourages the model to focus on plausible targets (e.g., faces) and avoid extreme failure cases. The expert is trained by minimizing the gaze loss $\mathcal{L}_{gaze}$ on this modified data:
\begin{equation}
  \underset{\theta_{face}}{\min} \sum_{I_i \in \mathcal{D}} \mathcal{L}_{gaze}(Ex_{aware}(\text{Aug}(I_i), B_{head}), \hat{M}_{gaze})
\end{equation}

\subparagraph{Socially Agnostic Gaze Expert ($Ex_{agnostic}$).}
Conversely, for the frequent non-social and unsure scenarios, the framework needs an expert that is maximally performant on those cases. The rationale is to let this model fully optimize for the most common situations without being constrained by the specialized requirements of the `Face' class. Thus, its parameters, $\theta_{agnostic}$, are optimized by minimizing the same gaze loss $\mathcal{L}_{gaze}$ on the original, unmodified training set $\mathcal{D}$:
\begin{equation}
   \underset{\theta_{agnostic}}{\min} \sum_{I_i \in \mathcal{D}} \mathcal{L}_{gaze}(Ex_{agnostic}(I_i, B_{head}), \hat{M}_{gaze})
\end{equation}

\begin{table*}[ht]
\centering
\footnotesize
\begin{tabular}{@{}lccccccccccc@{}}
\toprule
\textbf{Model} & \textbf{$L^2$↓} & \textbf{$L^2_{obj}$↓} & \textbf{$L^2_{face}$↓} & \textbf{$L^2_{pnf}$↓} & \textbf{Precision} & \textbf{Recall} & \textbf{F1} \\
\midrule
Qwen2.5-VL-7B-Instruct-AGT   & 0.0475 & 0.0413 & 0.1237 & 0.0837 & 0.7050 & 0.6260 & 0.6630 \\
\midrule
Sharingan                   & 0.0615 & 0.0615 & 0.0647 & 0.0561 & 0.4377 & 0.8010 & 0.5660 \\
Sharingan-AGT               & 0.0486 & 0.0451 & 0.0949 & 0.0595 & 0.7744 & 0.7330 & 0.7531 \\
Sharingan-SACF (Ours)        & \textbf{0.0480} & 0.0453 & 0.0817 & 0.0616 & 0.7647 & 0.7573 & \textbf{0.7610} \\
\midrule
Sharingan-SACF (Upper Bound)     & 0.0478 & 0.0471 & 0.0378 & 0.0923 & 0.9581 & 1.0000 & 0.9786 \\
\midrule
GazeLLE                     & 0.0670 & 0.0630 & 0.1092 & 0.1041 & 0.3868 & 0.6553 & 0.4865 \\
GazeLLE-AGT                  & 0.0460 & 0.0405 & 0.1130 & 0.0804 & 0.6984 & 0.6408 & 0.6684 \\
GazeLLE-SACF (Ours)         & \textbf{0.0453} & 0.0405 & 0.1019 & 0.0804 & 0.7041 & 0.6699 & \textbf{0.6866} \\
\midrule
GazeLLE-SACF (Upper Bound)     & 0.0396 & 0.0391 & 0.0307 & 0.0752 & 0.9856 & 0.9951 & 0.9903 \\
\bottomrule
\end{tabular}
\caption{Performance comparison of different gaze target detection methods. Metrics include various $L^2$, and standard classification scores.}
\label{tab:main_results}
\end{table*}

\paragraph{Gating and Inference.}
The final step is to integrate the coarse assessment with the fine-grained expert predictions during inference. Our rationale is that a decisive, dynamic routing mechanism is the most effective way to leverage the experts' specialized knowledge to predict gaze target coordinates:

\begin{equation}
    \mathbf{p}_t^* =
    \begin{cases}
        \underset{\mathbf{p}_t}{\arg\max}(Ex_{aware}((\text{Aug}(I), B_{head})) & \text{if } c_{coarse}=1 \\
        \underset{\mathbf{p}_t}{\arg\max}(Ex_{agnostic}(I, B_{head})) & \text{if } c_{coarse}=0
    \end{cases}
\end{equation}

While the coarse assessment guides the choice of expert, the final semantic classification, $c_{fine}$, is refined based on these precise output coordinates. Let $\mathcal{B}_{faces} = \{B_{face}^{(1)}, \dots, B_{face}^{(k)}\}$ be the set of all adult face bounding boxes in the image. The final class is determined by checking if the predicted point $\mathbf{p}_t$ falls within the union of these bounding box regions:
\begin{equation}
    c_{fine} =
    \begin{cases}
        \text{Face} & \text{if } \mathbf{p}_t \in \bigcup_{B \in \mathcal{B}_{faces}} \text{Area}(B) \\
        \text{Not face} & \text{otherwise}
    \end{cases}
\end{equation}
This two-step classification process—a coarse routing followed by a fine-grained spatial check—allows SACF to dynamically apply the correct specialist while making a final, evidence-based semantic judgment, effectively resolving the performance trade-offs imposed by the imbalanced data.

\section{Experiments}
\label{experiments}

\subsection{Experimental Setup}
\paragraph{Dataset.} We use our Autism Gaze Target (AGT) dataset with the splits described previously.

\paragraph{Models.} We compare our SACF against other models:
\begin{itemize}
    \item \textbf{Sharingan \cite{tafasca2024sharingan}:} A recent model trained on the Childplay dataset with neurotypical children.
    \item \textbf{GazeLLE \cite{ryan2025gaze}:} Current SOTA trained on the Childplay.
    \item \textbf{Sharingan-AGT:} Sharingan trained on our AGT dataset.
    \item \textbf{GazeLLE-AGT:} GazeLLE trained on our AGT dataset.
    \item \textbf{Qwen2.5-VL-7B-Instruct-AGT:} The MLLM fine-tuned on our AGT training set.
    \item \textbf{SACF (Ours):} Our Socially Aware Coarse-to-Fine gaze detection framework. We select two recent gaze target detection SOTA models: Sharingan and GazeLLE as the base models of our framework.
    \item \textbf{SACF w/ SCA upper-bound:} A version of our model that uses a perfect gate model (i.e., an oracle), showing the framework's potential with an ideal social context awareness module.
\end{itemize}

\paragraph{Evaluation Metrics.}
Following established practice in gaze target detection
\cite{nips15_recasens, chong2020detecting, fang2021dual, tafasca2023childplay, ryan2025gaze}, we report seven complementary scores that jointly capture
(i) \textit{where} the model points and
(ii) \textit{whether} it flags the scene as social (\emph{Face}) or not.

\begin{itemize}
  \item \textbf{$L^2$} - mean Euclidean distance (normalised) between the predicted and ground-truth gaze point over the test set.
  \item \textbf{$L^2_{obj}$} / \textbf{$L^2_{face}$} / \textbf{$L^2_{pnf}$} - the same error computed only on frames whose ground-truth target is, respectively, an \emph{Object}, an \emph{Adult Face}, or a \emph{Person-Non-Face} region (e.g.\ torso or hands).  Using three separate classes makes minority-class performance visible instead of being averaged away.
  \item \textbf{Precision}, \textbf{Recall}, \textbf{F1} - macro-averaged over the binary decision \emph{Face} vs.\ \emph{Not-face}.  Macro averaging prevents the 93.4 \% majority class from inflating the score.
\end{itemize}

\paragraph{Implementation Details.}
All models were implemented in PyTorch. We used a Qwen2.5VL-7B as the gate model and GazeLLE for our expert models. Training and testing were performed using NVIDIA A6000 GPUs.

\begin{table}[t]
\centering
\begin{tabular}{@{}lcccc@{}}
\toprule
\textbf{Class} & \textbf{Precision} & \textbf{Recall} & \textbf{F1 Score} & \textbf{Support} \\
\midrule
Face & 69.23\% & 65.53\% & 0.673 & 206 \\
Not-face & 97.76\% & 98.10\% & 0.979 & 3,158 \\
\bottomrule
\end{tabular}
\caption{SAC's performance for the Face vs. Not-face task.}
\label{tab:classification_report}
\end{table}

\begin{table*}[ht]
\centering
\footnotesize
\begin{tabular}{@{}l r c c l@{}}
\toprule
\textbf{Scenario} & {Sample Count} & {$Ex_{aware}$'s $L^2$} & {$Ex_{agnostic}$'s $L^2$} & {Winner} \\
\midrule
\multicolumn{5}{l}{\textit{Correct Predictions}} \\
Pred=Face, GT=Face & 135 & 0.0303 & 0.0898 & $Ex_{aware}$ (3x better) \\
Pred=Not-face, GT=Not-face & 3,098 & 0.0398 & 0.0409 & $Ex_{aware}$ (slightly better) \\
\midrule
\multicolumn{5}{l}{\textit{Incorrect Predictions}} \\
Pred=Face, GT=Not-face & 60 & 0.2818 & 0.0817 & $Ex_{agnostic}$ (3.4x better) \\
Pred=Not-face, GT=Face & 71 & 0.2090 & 0.1569 & $Ex_{agnostic}$ (1.3x better) \\
\bottomrule
\end{tabular}
\caption{Analysis of expert models' performance for different prediction scenarios. The table is grouped by correct and incorrect predictions. As we can see, since misclassifications can hurt overall performance, it is essential to obtain a good SCA.}
\label{tab:expert_analysis}
\end{table*}

\subsection{Results and Analysis}\label{sec:results_analysis}

\noindent
\textbf{Overall gains across the board.} Table~\ref{tab:main_results} shows that our \textbf{SACF} framework achieves new state-of-the-art results.  Using the \textit{Sharingan} backbone, SACF reduces the $L^2$ from 0.0486 to 0.0480, while pushing the F1 score from 0.7531 to 0.7610. A similar trend is observed with the \textit{GazeLLE} backbone, where overall $L^2$ drops from 0.0460 to 0.0453 and F1 climbs from 0.6684 to 0.6866.

\begin{figure}[t]
\centering
\includegraphics[width=\columnwidth]{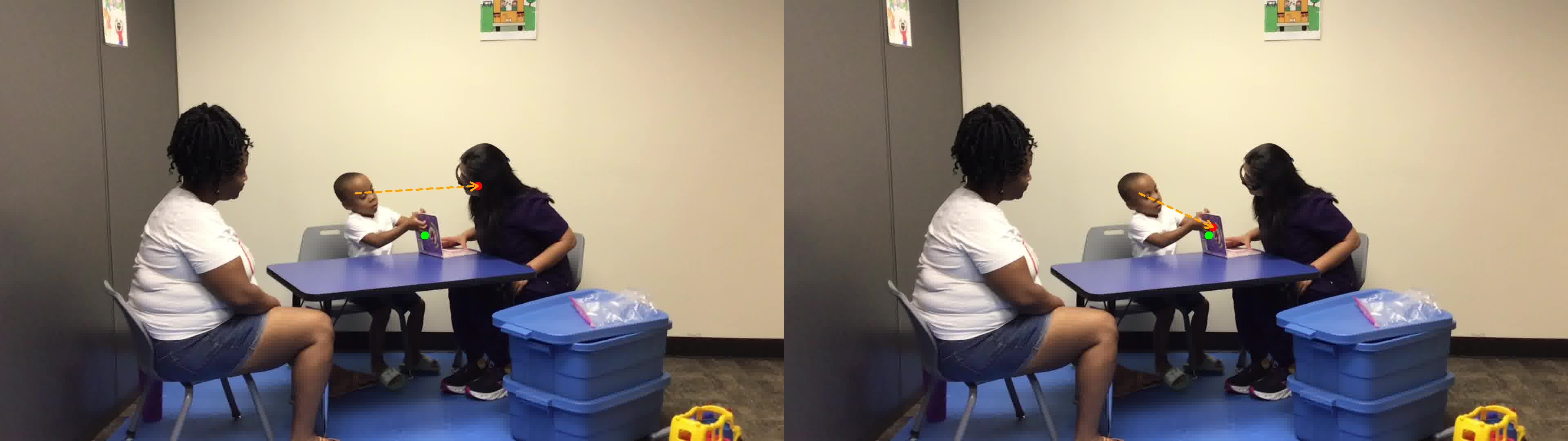}
\caption{A child is playing with an object. However, the baseline Sharingan model trained on the neurotypical Childplay dataset (left) is biased and predicts the child is looking at the clinician's face. When trained on our AGT dataset (right), it correctly predicts the child is looking at the object. The green dot denotes the ground-truth gaze target, while the red dot denotes the predicted target.}
\label{fig:qualitative_results_sharingan}
\end{figure}

\noindent
\textbf{Models trained on neurotypical data fail on our autism dataset.} We found that models trained on neurotypical datasets, like the original Sharingan and GazeLLE, perform poorly on our AGT test set (Table~\ref{tab:main_results}) by overestimating the social engagement of autistic children. Their performance improves significantly after being trained on our AGT dataset (Figure~\ref{fig:qualitative_results_sharingan}), highlighting the social distribution shift and the need for autism-specific data.

\begin{figure}[t]
\centering
\includegraphics[width=\columnwidth]{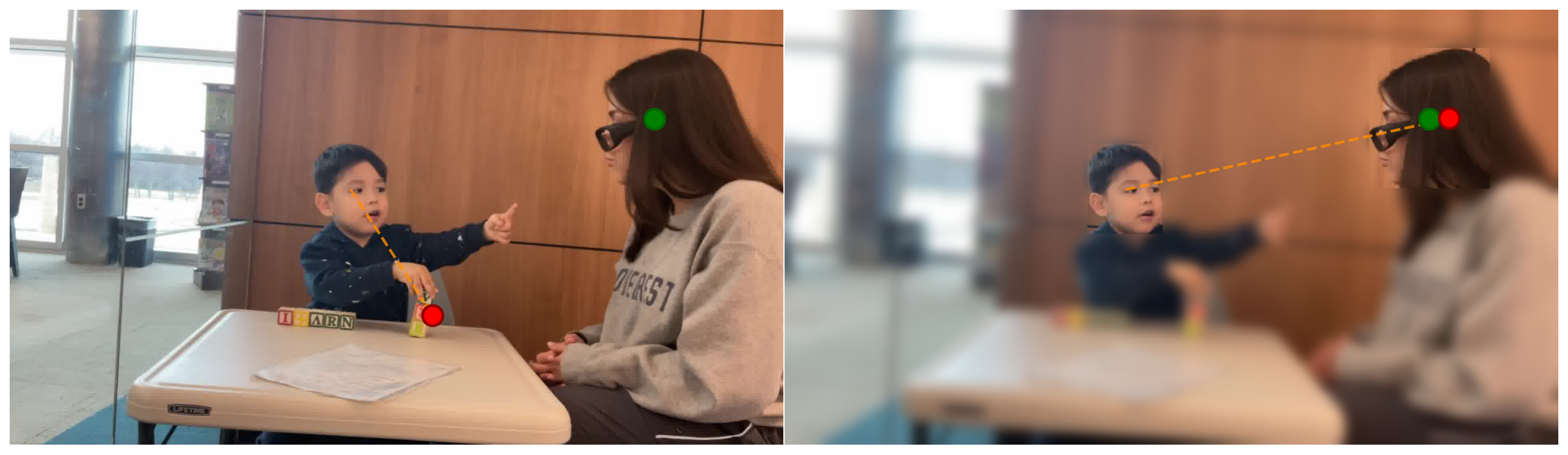}
\caption{A child is actively interacting with a clinician. Our SACF framework (right) correctly localizes gaze to the face, thanks to the prior knowledge about social context leveraged from the MLLM. The simply trained GazeLLE-AGT (left) fails due to a lack of this social context awareness. Green dot denotes the ground-truth, while red denotes the prediction.}
\label{fig:qualitative_results}
\end{figure}

\noindent
\textbf{Excelling on the rare but crucial \textit{Face} class.} Because only 6.6\% of gaze events are directed at a face (Table~\ref{tab:dataset_stats}), improvements on this minority class are easily masked by majority performance. SACF overturns that bias: on \textit{GazeLLE}, \textbf{$L^2_{face}$} falls from 0.1130 to 0.1019;($\downarrow$9.8\%), and on \textit{Sharingan} from 0.0949 to 0.0817;($\downarrow$13.9\%). This directly translates into more reliable detection of socially relevant moments in clinical practice (Figure \ref{fig:qualitative_results}).

\noindent
\textbf{An expert-level view of why SACF works.} Table~\ref{tab:expert_analysis} reveals that the Socially Aware Gaze Expert is 3$\times$ more accurate than the Socially Agnostic Gaze Expert when a face is indeed the target. Conversely, when the scene is mis-routed, the Socially Agnostic Gaze Expert is up to 3.4$\times$ better. The Context-Aware gate therefore acts as a high-stakes switch: when it chooses correctly (96\% of the time), the framework benefits from the best of both specialists; when it errs, performance degrades gracefully rather than catastrophically.

\noindent
\textbf{Upper-bound analysis highlights room for growth.} With the performance of our current small MLLM SCA (Table~\ref{tab:classification_report}), we are already able to improve the final performance of the framework by routing input to different experts based on social context. Replacing the learned gate with an oracle (\textit{SACF Upper Bound}) yields an F1 of 0.9903 and an almost perfect \textit{$L^2_{face}$} of 0.0307. This gap pinpoints the gate as a promising lever for future gains: any advancement in social-context recognition will directly unlock further accuracy. Considering the current rapid development of MLLMs, our framework also stands to benefit significantly.

\noindent
\textbf{Clinical impact.} The 10-14\% drop in Face-target localization error means that, on average, the predicted gaze point lands 4-6 pixels closer to the true facial region at a 224$\times$224 resolution. In downstream joint-attention related scoring pipelines, this improves F1 by around 2\% (Table~\ref{tab:main_results}), bringing automated assessments one step closer to human-expert reliability.

\section{Conclusion}
\label{conclusion}
In this paper, we introduced gaze target detection in autistic children, a critical task for developing AI-powered tools to support the autism community. We addressed the scarcity of appropriate data by collecting and annotating the AGT dataset, which consists of 16,582 frames of young autistic children in naturalistic play sessions. The dataset's inherent class imbalance, a direct result of the behavioral characteristics of autism, renders existing gaze detection methods inadequate.
To overcome this challenge, we proposed the Socially Aware Coarse-to-Fine (SACF) gaze detection framework. By first assessing the social context of a scene with Social Context Awareness (SCA) module and then using a gated two-pathway system of experts, our model achieves new state-of-the-art performance. It significantly improves the detection of rare but clinically vital gaze-to-face events without compromising its accuracy on more frequent non-social targets.
Our work represents a substantial step towards building automated systems for objective, scalable measurement of gaze targets for young autistic children.

\section*{Acknowledgments}
This work was supported by a UT Dallas Seed Program for Interdisciplinary Research (SPIRe) Award and grants from the Texas Higher Education Coordinating Board (THECB) Autism Grants Program (Rollins, PI; NOGA \#20476 \& \#27509). The opinions and conclusions expressed are those of the authors and do not necessarily represent the opinions or policies of the funding sources.

\bibliography{aaai2026}

\end{document}